\documentclass[final]{cvpr}

\usepackage{times}
\usepackage{epsfig}
\usepackage{graphicx}
\usepackage{amsmath}
\usepackage{amssymb}
\usepackage{algorithm}
\usepackage[noend]{algpseudocode}
\usepackage{booktabs}
\usepackage{color}
\usepackage{color, colortbl}
\definecolor{Gray}{gray}{0.9}
\usepackage{pifont}
\newcommand{\cmark}{\ding{51}}%
\newcommand{\xmark}{\ding{55}}%

\usepackage{color,soul}
\definecolor{lightcornflowerblue}{rgb}{0.6, 0.81, 0.93}

\usepackage[pagebackref=true,breaklinks=true,colorlinks,bookmarks=false]{hyperref}

\newcommand{\fs}[1]{\textcolor{black}{#1}}

\definecolor{sgclr}{rgb}{1.0, 0.0, 1.0}

\def\cvprPaperID{} 
\def\confYear{}

\begin{document}

\title{Probabilistic Tracklet Scoring and Inpainting for Multiple Object Tracking}

\author{
Fatemeh Saleh$^{1,2}$, Sadegh Aliakbarian$^{1,2}$, Hamid Rezatofighi$^{3}$, Mathieu Salzmann$^{4}$, Stephen Gould$^{1,2}$\\
$^{1}$Australian National University, $^{2}$ACRV, $^{3}$Monash University, $^{4}$CVLab-EPFL\\
{\small \tt fatemehsadat.saleh@anu.edu.au}
}

\maketitle

\begin{abstract}
   Despite the recent advances in multiple object tracking (MOT), achieved by joint detection and tracking, dealing with long occlusions remains a challenge. This is due to the fact that such techniques tend to ignore the long-term motion information. In this paper, we introduce a probabilistic autoregressive motion model to score tracklet proposals by directly measuring their likelihood. This is achieved by training our model to learn the underlying distribution of natural tracklets. As such, our model allows us not only to assign new detections to existing tracklets, but also to inpaint a tracklet when an object has been lost for a long time, e.g., due to occlusion, by sampling tracklets so as to fill the gap caused by misdetections. Our experiments demonstrate the superiority of our approach at tracking objects in challenging sequences; it outperforms the state of the art in most standard MOT metrics on multiple MOT benchmark datasets, including MOT16, MOT17, and MOT20.
\end{abstract}

\section{Introduction}
\label{sec:introduction}
Tracking multiple objects in a video is key to the success of many computer vision applications, such as sport analysis, autonomous driving, robot navigation, and visual surveillance. 
With the recent progress in object detection, tracking-by-detection~\cite{andriluka2008people} has become the de facto approach to multiple object tracking; it consists of first detecting the objects in the individual frames and then associating these detections with trajectories, known as tracklets. 
While these two steps were originally performed sequentially, recent advances have 
benefited from treating detection and tracking jointly~\cite{bergmann2019tracking, tai2020chained,zhou2020tracking}.
These approaches cast MOT as a \emph{local tracking} problem, utilizing either an object detector's regression head~\cite{bergmann2019tracking} or an additional offset head~\cite{tai2020chained,zhou2020tracking} to perform temporal re-alignment of the object bounding boxes in consecutive frames. In other words, these approaches treat tracking as the problem of propagating detection identities across consecutive frames.
While this strategy constitutes the state of the art on many benchmark datasets in terms of MOT metrics that highlight the quality of the detections, \textit{e.g.}, MOTA, it fails to maintain identities throughout occlusions, and thus tends to produce many identity switches. 
In this paper, we 
address this issue
by developing a stochastic motion model that helps the tracker to maintain identities, even in the presence of long-term occlusions. In other words, we show that, while largely ignored in the recent MOT literature, motion remains a critical cue for tracking, even with the great progress achieved by detectors. This is evidenced by our experimental results on multiple MOT benchmark datasets, in which our approach outperforms the state of the art by a large margin.

Motion has, of course, been considered in the past, mostly in the tracking-by-detection literature, via either 
model-based filtering techniques~\cite{bewley2016simple,kalman1960new,Wojke2017simple} or more sophisticated data-driven
ones based on RNNs~\cite{dicle2013way, fang2018recurrent,milan2017online,ran2019robust, sadeghian2017tracking,wan2018online,yang2012multi,yang2012online}. However, all of these approaches treat human motion as a deterministic or a uni-modal process. Here, we argue that human motion is a stochastic multi-modal process, and should thus be modeled stochastically. 
Note that a similar concept has also been explored in the context of trajectory forecasting, where the problem is to often given perfect (ground-truth) trajectories,  predict fixed-length continuations of those trajectories as a single path~\cite{alahi2016social,huang2019stgat}, or a distribution over different paths~\cite{gupta2018social,ivanovic2019trajectron,kosaraju2019social,mangalam2020pecnet,sadeghian2019sophie,SalzmannIvanovicEtAl2020}. However, to the best of our knowledge, these techniques have not been incorporated in the context of MOT, where we deal with noisy observations (detections), frequent occlusions, and assignment uncertainties.

Therefore, we introduce a 
stochastic
autoregressive motion model that explicitly learns the multi-modal distribution of natural trajectories. This allows us to estimate the likelihood of a tracklet given a sequence of bounding box locations and the tracklets of the surrounding agents. We then use this model to compute the likelihood of a tracklet after assigning it a new detection.
Moreover, learning the multi-modal distribution of tracklets allows us to inpaint a tracklet in the presence of misdetections caused by occlusion by sampling from the learned distribution. This is also what the visual cortex of the human brain does when reasoning about dynamically occluded objects~\cite{erlikhman2017decoding, slotnick2005visual}.

To summarize, our contributions are as follows:
    \textbf{(1)}~We introduce a stochastic autoregressive model to score a tracklet by the likelihood that it represents natural motion.
    \textbf{(2)} Since our model learns the multi-modal distribution of natural human motion, it can generate multiple plausible continuations of the tracklets and inpaint tracklets containing missed detections.
    \textbf{(3)} 
    Our stochastic motion model can better preserve identities over longer time horizons than recent MOT approaches, especially when there are occlusions.
   
   We conduct comprehensive ablation studies, demonstrating the effectiveness of the different components of our approach. Our method outperforms the state of the art in multiple MOT benchmark datasets, particularly improving the metrics related to long-term identity preservation, such as IDF1, ID Switch (IDs), and Mostly Tracked Tracklets (MT). This is further confirmed by our experiments on the challenging new MOT20~\cite{dendorfer2020mot20} dataset, targeting highly crowded scenarios.
We refer to our model as \textbf{ArTIST}, for \textbf{A}uto\textbf{r}egressive \textbf{T}racklet \textbf{I}npainting and \textbf{S}coring for \textbf{T}racking. 

\section{Related Work}
\label{sec:related_work}
\paragraph{Tracking-by-Detection.}
Tracking-by-detection~\cite{andriluka2008people} has proven to be effective to address the MOT problem.
In this context, tracking systems can be roughly grouped into online ones~\cite{bergmann2019tracking,chu2019online,chu2019famnet,chu2017online,leal2016learning,long2018real,maksai2017non,milan2017online,sadeghian2017tracking,xu2019spatial,yoon2018online,zhu2018online}, where the tracklets are grown at each time step, and batch-based (or offline) ones~\cite{braso2020learning,kim2015multiple,kim2018multi,maksai2019eliminating,tang2017multiple,yoon2018multiple}, where the tracklets are computed after processing the entire sequence, usually in a multiple hypothesis tracking (MHT) framework~\cite{blackman2004multiple,kim2015multiple}. 
In this paper, we develop an online tracking system and thus, in this section, focus on this class of methods.

Closest to our approach are the ones that design or utilize a motion model for state prediction. 
In~\cite{bewley2016simple, hamid2015joint,Wojke2017simple}, this was achieved with a Kalman Filter~\cite{kalman1960new} aiming to approximate the inter-frame displacements of each object with a linear constant velocity model, assuming independence across the objects and from the camera motion.  As a linear motion model often poorly reflects reality, more sophisticated data-driven motion models have been proposed to permit more complex state prediction~\cite{dicle2013way,fang2018recurrent,milan2017online,ran2019robust,sadeghian2017tracking,wan2018online,yang2012multi,yang2012online}. In particular, the use of recurrent networks was introduced in~\cite{milan2017online} to mimic the behavior of a Bayesian filter for motion modeling and associations. 
Following~\cite{milan2017online}, several recurrent approaches have been developed for MOT.  In~\cite{fang2018recurrent,liang2018lstm,ran2019robust,sadeghian2017tracking}, multiple streams of RNNs have been utilized to incorporate different forms of information, such as appearance and motion, to compute a score for the assignment, usually done by solving an assignment problem via the Munkres (a.k.a. Hungarian) algorithm~\cite{munkres1957algorithms}.

\begin{figure*}[!ht]
    \centering
    \includegraphics[width=0.9\textwidth]{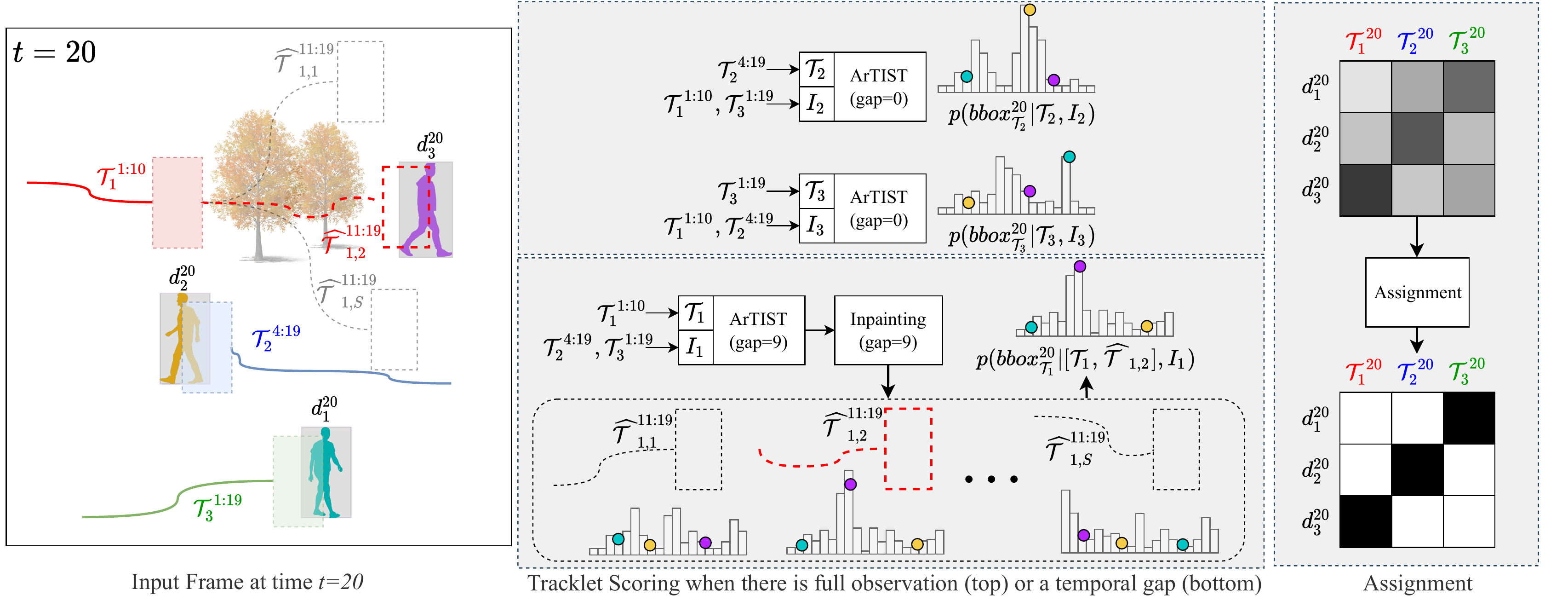}
    \caption{An overview of our framework. \textbf{Left}: At time $t=20$, we are provided with detections illustrated by solid gray boxes and a set of alive tracklets $\mathcal{T}_{\cdot}^{\cdot}$, shown in different colors. The task is to assign the new detections to existing tracklets. \textbf{Middle}: An illustration of tracklet scoring (top) in the case of full observation and tracklet inpainting and scoring (bottom) in case of misdetection due to \textit{e.g.}, occlusion. \textbf{Right}: Given the computed scores, we solve the Munkres algorithm to perform assingment before processing the next frame.}
    \label{fig:artist_arch}
\end{figure*}

In all the aforementioned approaches, human motion has been treated either deterministically~\cite{ran2019robust,sadeghian2017tracking} or probabilistically in a uni-modal manner~\cite{fang2018recurrent,wan2018online}.
The shortcoming of such techniques is that, while these are reasonable design choices when the state estimation uncertainty is low, they become poorly suited for tracking throughout long occlusions, where uncertainty increases significantly.
This is particularly due to the stochastic nature of human motion, a property that has been overlooked by existing approaches. 
\\
\noindent\textbf{Joint Detection and Tracking.}
As an alternative to the two-stage tracking-by-detection, the recent trend in MOT has moved toward jointly performing detection and tracking. This is achieved by converting an object detector to predict the position of an object in the next frame, thus inherently utilizing it for tracking.
To this end, Tracktor (and its variants, Tracktor++ and Tracktor++v2)~\cite{bergmann2019tracking}  exploits the regression head of a Faster R-CNN~\cite{ren2015faster} to perform temporal realignment of the object bounding boxes. 
CenterTrack~\cite{zhou2020tracking} adapts the CenterNet object detector~\cite{zhou2019objects} to take two frames and a heatmap rendered from the tracked object centers as input, and computes detection and tracking offsets for the current frame. Chained-Tracker~\cite{tai2020chained} uses two adjacent frames as input to regress a pair of bounding boxes for the same target in the two adjacent frames. 
Although these approaches yield impressive results, 
their effectiveness depends on the feasibility of detecting the objects. In fact, these approaches look at the tracking problem from a \textit{local} perspective, and thus, even though they use techniques such as person ReID~\cite{bergmann2019tracking,hermans2017defense}, CMC~\cite{bergmann2019tracking,evangelidis2008parametric}, or ReBirth~\cite{bergmann2019tracking,zhou2020tracking} to re-identify occluded objects, tend to struggle to preserve identities.
\\
\noindent\textbf{This paper.} To address the shortcomings of the aforementioned approaches, we introduce a MOT framework with a focus on designing a non-linear \emph{stochastic} motion model by learning the multi-modal distribution of the next plausible states of a pedestrian so as to reason about uncertainties in the scene when facing occlusions. It not only allows us to estimate the likelihood of a tracklet and directly use it for scoring a new detection, but also enables us to fill in the gaps in case of misdetection caused by occlusion by sampling from the learned distribution. As a result, we considerably improve identity preservation, as confirmed by our results on several MOT benchmark datasets.

\section{Proposed Method}
\label{sec:proposed_method}
We address the problem of online tracking of multiple objects in a scene by designing a stochastic 
motion model. 
In this section, we first define our notation, and then provide an overview of our ArTIST algorithm, followed by the details of its different components.

\subsection{Notation}
\label{sec:notation}
As many other online tracking systems, we follow a  tracking-by-detection paradigm~\cite{andriluka2008people}. 
Let us consider a video of $T$ frames, where, for each frame, we are provided with a set of detections computed by, \textit{e.g.}, Faster-RCNN~\cite{ren2015faster}. This yields an overall detection set for the entire video denoted by $\mathcal{D}^{1:T} = \{D^1, D^2, ..., D^T\}$, where $D^t=\{d_1^t, d_2^t,...\}$ is the set of all detections at time $t$, with $d_i^t\in\mathbb{R}^4$, \textit{i.e.}, the 2D coordinates $(x,y)$ of the top-left bounding box corner, its width $w$ and height $h$. 
We tentatively initialize a first set of tracklets $\mathbb{T}$ with the detections $D^1$ in the first frame. From the second time-step to the end of the video, the goal is to expand the tracklets by assigning the new detections to their corresponding tracklets. 
Throughout the video, new tracklets may be created, and incorporated into the set of tracklets $\mathbb{T}$, and existing tracklets may be terminated and removed from $\mathbb{T}$. We write $\mathbb{T}=\{\mathcal{T}_1^{{s_1}:{e_1}}, \mathcal{T}_2^{{s_2}:{e_2}}, ..., \mathcal{T}_m^{{s_m}:{e_m}}\}$, where $\mathcal{T}_j^{{s_j}:{e_j}}$ is a tracklet representing the $j^{th}$ identity that has been alive from time ${s_j}$ to time ${e_j}$, and is defined as $\mathcal{T}_j^{{s_j}:{e_j}}=\{d_{\Pi_j}^{{s_j}}, d_{\Pi_j}^{{s_j}+1}, ..., d_{\Pi_j}^{{e_j}}\}$, where $d_{\Pi_j}^{t}$ 
is the detection (or an inpainted box) at time $t$ that has been assigned to tracklet $\mathcal{T}_j^{{s_j}:{e_j}}$. 
For each tracklet $\mathcal{T}_j^{{s_j}:{e_j}}$, 
we define a learnable interaction representation $I_j^{s_j:e_j}$ which captures the latent representations of all other tracklets whose lifespan overlaps with the temporal range $[{{s_j}, {e_j}}]$. We also define $z^{t}_j$ that captures the hidden representation of $\mathcal{T}_j^{{s_j}:{t}}$. Both $I_j$ and $z_j$ are described in detail below.

\subsection{ArTIST Overview}
\label{sec:overview}

ArTIST relies on two main steps for every video frame: scoring how well a detection fits in an existing tracklet (as in Fig.~\ref{fig:artist_arch}-middle) and assigning the detections to the tracklets 
(as in Fig.~\ref{fig:artist_arch}-right), thus updating them.

\begin{figure*}
    \centering
    \includegraphics[width=.9\textwidth]{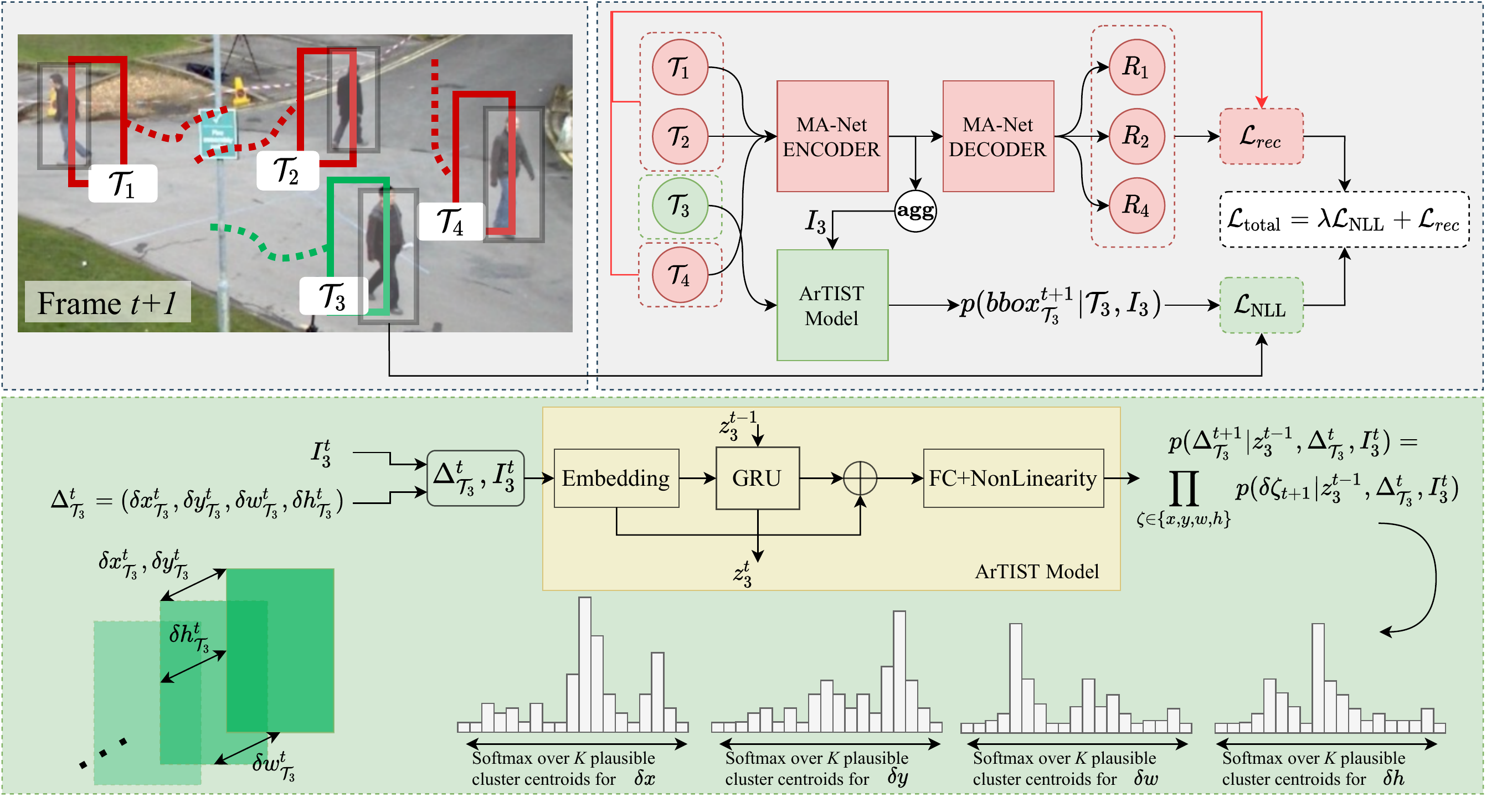}
    \caption{\textbf{Top}: An overview of our approach during training. Given a set of GT bounding boxes (gray boxes) at time $t+1$, we show the training procedure of our model that aims to maximize the likelihood of tracklet $\mathcal{T}_3$ when assigned with the correct detection by incorporating additional information from all other tracklets, $I_3$. MA-Net autoencoder is trained jointly with the ArTIST model to provide an expressive representation of $I_3$ by learning to reconstruct other tracklets (shown in red). \textbf{Bottom}: An overview of the recurrent residual architecture of ArTIST model for processing a tracklet at time $t$ to compute the probability distribution of the next plausible bounding box. Such distribution is used to either evaluate the likelihood of assigning a new detection to the tracklet or for inpainting a tracklet.}
    \label{fig:artist_train_test}
\end{figure*}

Specifically, given an input frame at time $t$, \textit{e.g.}, $t=20$ in Fig.~\ref{fig:artist_arch}-left, a set of 
tracklets 
up to time $t-1$, \textit{e.g.}, $\mathcal{T}_1^{1:10}$, $\mathcal{T}_2^{4:19}$, and $\mathcal{T}_3^{1:19}$, and a set of detections at time $t$, \textit{e.g.}, $d_1^{20}$, $d_2^{20}$, and $d_3^{20}$, shown as solid gray boxes, we perform 
scoring for the tracklets were last assigned a detection at time $t-1$, \textit{i.e.}, the non-occluded tracklets.
This is denoted by $\text{gap}=0$ in Fig.~\ref{fig:artist_arch}-middle.
We refer to these tracklets as {\it alive}, and to others as {\it tentatively alive}.
For each alive tracklet, for instance $\mathcal{T}_2^{4:19}$, ArTIST computes the probability distribution of the next plausible bounding box ($bbox_{\mathcal{T}_2}^{20}$) that can be assigned to  $\mathcal{T}_2^{4:19}$, given information about both this tracklet and the other tracklets that interact with it, \textit{i.e.}, $\mathcal{T}_2$ and $I_2$, respectively. We then evaluate all the detections $d_i^t \in D^t$ at time $t$ under the estimated distribution.

For any tentatively alive tracklets (\textit{e.g.}, $\mathcal{T}_1^{1:10}$), whose last assignment occurred prior to $t-1$, resulting in a non-zero gap, we first perform tracklet inpainting to fill the gap up to $t-1$, so that it can be considered as a fully-observed tracklet. As ArTIST estimates a multi-modal distribution of natural motion, we generate $S$ plausible tracklets to fill in this gap, denoted by $\{\widehat{\mathcal{T}}_{1,1}^{11:19}, ..., \widehat{\mathcal{T}}_{1,S}^{11:19}\}$ in the bottom part of Fig.~\ref{fig:artist_arch}-middle. We then select the best inpainted tracklet (the second one in this example) among the $S$ candidates to complete $\mathcal{T}_1^{1:19}$ as $[\mathcal{T}_1^{1:10}, \widehat{\mathcal{T}}_{1,2}^{11:19}]$. 
We can now treat this tracklet as having zero gap and thus compute the distribution over next plausible bounding box assignments. 

Finally, as illustrated in Fig.~\ref{fig:artist_arch}-right, we construct a cost matrix from the likelihoods of each detection under the estimated distributions for all tracklets, and obtain an optimal assignment using the Munkres algorithm~\cite{munkres1957algorithms}. We then update all the tracklets with the assigned detections, and repeat the entire process for the next time-step. In the following sections, we provide more details about our ArTIST architecture and the different steps of this algorithm, with a complete specification given in the supplementary material.

\subsection{ArTIST Architecture}
\label{subsec:artist_arch}
ArTIST is a stochastic autoregressive motion model that aims to explicitly learn the distribution of natural tracklets. As an estimator, ArTIST is capable of determining the likelihood of each tracklet. As a generative model, ArTIST is capable of generating multiple plausible continuations of a tracklet by multinomial sampling from the estimated multi-modal distribution at each time-step. 

The probability of a tracklet $\mathcal{T}_j^{{s_j}:t}$, where $t$ is the current time-step, in an autoregressive framework is defined as 
\begin{align}
    p(\mathcal{T}_j^{{s_j}:t}|I_j^{{s_j}:t}) = p(d^{{s_j}}_{\Pi_j}|I_j^{{s_j}})\!\!\!\prod_{k={s_j}+1}^{t}\!\!\!p(d^k_{\Pi_j} \mid d^{<k}_{\Pi_j}, I^{<k}_j)\,,
    \label{eq:autoregressive}
\end{align}
where $d^{k}_{\Pi_j}$ is the detection assigned to $\mathcal{T}_j$ at time $k$ and $I^k_j$ is the representation of the interactions computed from other tracklets co-occurring with $\mathcal{T}_j$ at time $k$. 
Since each detection is represented by continuous bounding box coordinates, one could attempt to regress its position in the next frame given previous positions.
However, regression does not explicitly provide a distribution over natural tracklets. Furthermore, regression can only generate a single deterministic continuation of a tracklet, which does not reflect the stochastic nature of, \textit{e.g.}, human motion, for which multiple continuations may be equally likely. 
 
To remedy this, inspired by PixelRNN~\cite{oord2016pixel}, we propose to discretize the bounding box position space. This allows us to model $p(\mathcal{T}_j^{{s_j}:{e_j}})$ as a discrete distribution, with every conditional distribution in Eq.~\ref{eq:autoregressive} modeled as a multinomial (categorical) distribution with a \textit{softmax} layer. However, unlike PixelRNN-like generative models that discretize the space by data-independent quantization, \textit{e.g.}, through binning, we define a data-dependent set of discrete values by clustering the motion velocities, \textit{i.e.}, $\delta x$, $\delta y$, $\delta w$, and $\delta h$, between consecutive frames, normalized by the width and height of the corresponding frames. This makes our output space shift and scale invariant.
In practice, we use non-parametric k-means clustering~\cite{macqueen1967some} to obtain $K$ clusters, and treat each cluster centroid as a discrete motion class.

Our ArTIST architecture is depicted by Fig.~\ref{fig:artist_train_test}, whose top portion illustrates the high-level overview of ArTIST during training. In general, during training the model takes as input all alive tracklets $\mathbb{T}$, and jointly learns the distribution of each tracklet, shown in green in Fig.~\ref{fig:artist_train_test}-top, together with a representation of the interactions, shown in red. 
Since we aim to estimate a probability distribution over the bounding box position in the next time-step, we train our model with a negative log-likelihood loss function. Additionally, to learn an expressive representation of the interactions, we use a moving agent autoencoder network (MA-Net) that is trained to reconstruct all the interacting tracklets, as discussed in more detail below. Thus, to train our model, we minimize 
\begin{align}
    \mathcal{L}_{\text{total}} = \lambda \mathcal{L}_{\text{NLL}} + \mathcal{L}_{rec}\;,
\end{align}
where $\mathcal{L}_{rec}$ is the mean squared error loss and $\lambda$ is an annealing function. We start from $\lambda=0$, forcing the model to learn better interaction representations first, and gradually increase it to $\lambda=1$, following a logistic curve, to account for both terms equally.

As shown in Fig.~\ref{fig:artist_train_test}-bottom, ArTIST itself relies on a recurrent residual architecture to represent motion velocities. At each time-step $t$, it takes as input a motion velocity represented by $\Delta_{\mathcal{T}_j}^t = (\delta x_{\mathcal{T}_j}^t, \delta y_{\mathcal{T}_j}^t, \delta w_{\mathcal{T}_j}^t, \delta h_{\mathcal{T}_j}^t)$ and an interaction representation $I_j^t$, discussed below. Given these inputs and the hidden state computed in the last time-step $z^{t-1}_j$, it predicts a distribution over the motion velocity for time $t+1$, \textit{i.e.}, $p(\Delta^{t+1}_{\mathcal{T}_j}\mid z^{t-1}_j, \Delta^t_{\mathcal{T}_j}, I^t_j)$.
This approximates the definition in Eq.~\ref{eq:autoregressive}, since $z^{t-1}_j$ carries information about all previous time-steps.
\\
\noindent{\textbf{Moving Agent Interactions.}}
Most of the existing MOT frameworks~\cite{bewley2016simple,fang2018recurrent,milan2017online,Wojke2017simple} treat each 
agent as independent from other agents in the scene.
A few approaches~\cite{maksai2019eliminating,sadeghian2017tracking} have nonetheless shown the benefits of modeling the interactions between agents. 
We believe that an effective modeling of interactions will lead to better tracking quality as the motion of each pedestrian may be affected by the behaviour of the other agents in the scene.
In this paper, we do so using the 
Moving Agent Network, MA-Net, illustrated in Fig.~\ref{fig:artist_train_test}. MA-Net is a recurrent autoencoder neural network that learns to reconstruct the tracklets of all moving agents potentially interacting with the tracklet of interest, \textit{e.g.}, $\mathcal{T}_j$. During training, the encoder compresses the tracklets into a latent representation (\textit{i.e.}, the hidden state of the last time-step), and the decoder reconstructs all tracklets given their compressed latent representations. 
To learn distribution of $\mathcal{T}_j$, ArTIST then needs a representation of the interacting agents that depends neither on their number nor on their order. We achieve this via max-aggregation of the latent representations of all interacting agents, $\mathbb{T}\setminus\{\mathcal{T}_j\}$. Specifically, we take the hidden-state of the last recurrent cell in the MA-Net encoder for the $\text{N}_{I_j}$ interacting agents, leading to a matrix in $\mathbb{R}^{\text{N}_{I_j}\times L}$, where $L$ is the hidden state dimension. We then perform max-pooling over the first dimension of this matrix, giving us $I_j\in \mathbb{R}^{L}$. 
Note that, during tracking (\textit{i.e.}, at test time), we remove the MA-Net decoder and only rely on the representation computed by its encoder.

\subsection{Tracklet Scoring}
\label{sec:scoring}
Given the trained ArTIST model, we can score how likely a detection at time $t$ is to be the continuation of a tracklet $\mathcal{T}_j$.
To this end, given $\mathcal{T}_j$'s velocity sequence and $I_j$, the model estimates a probability distribution over the location of the bounding box at time $t$. We then take the likelihood of the observed detection given the estimated distribution as a score for the tracklet-detection pair. Specifically, we compute the $\Delta$, \textit{i.e.}, the potential velocity of changes in $x,y,w,$ and $h$ made by
any detection with respect to the previous observation (or inpainted bounding box if the previous time-step was inpainted). We then take the probability estimated for the centroid closest to this $\Delta$ as likelihood. In practice, we assume independence of the bounding box parameters, \textit{i.e.}, $\delta x_{\mathcal{T}_j}^t$, $\delta y_{\mathcal{T}_j}^t$, $\delta w_{\mathcal{T}_j}^t$, and $\delta h_{\mathcal{T}_j}^t$. Therefore, we have four sets of clusters and thus four probability distributions estimated at each time-step, as shown in Fig.~\ref{fig:artist_train_test}-bottom. We then compute the likelihood of a bounding box as the product of the probabilities of the components, as
\begin{align}
    p(\Delta_{\mathcal{T}_j}^{t+1} \mid z^{t-1}_j, \Delta_{\mathcal{T}_j}^{t}, I_j^t) &= \!\!\!\!\!\!\prod_{\xi \in \{x, y, w, h\}}\!\!\!\!\!\!\! p(\delta \xi^{t+1}_{\mathcal{T}_j} \mid z^{t-1}_j, \Delta_{\mathcal{T}_j}^{t}, I_j^t).
\end{align}
In practice, we do this in log space, summing over the log of the probabilities.

\subsection{Tracklet Inpainting}
\label{sec:inpainting}
In the real world, detection failures for a few frames are quite common due to, \textit{e.g.}, occlusion. Such failures complicate the association of upcoming detections with the tracklets, and thus may lead to erroneous tracklet terminations. Our approach overcomes this by inpainting the tracklets for which no detections are available. Let us consider the scenario where a tracklet was not assigned any detection in the past few frames. We now seek to check whether a new detection at the current time-step belongs to it. To compute a likelihood for the new observation, we need to have access to the full bounding box sequence up to the previous time-step. To this end, we use our model to inpaint the missing observations, as illustrated in the bottom of the Fig.~\ref{fig:artist_arch}-middle, by \emph{multinomial} sampling from the learned tracklet distribution. Sampling can in fact be done autoregressively to create a diverse set of full sequence of observations and inpainted boxes, which, in turn, allows us to score a new detection. 
To account for the fact that motion is stochastic by nature, especially for humans, we sample $S$ candidates for the whole subsequence to inpaint from the estimated distribution and get multiple plausible inpainted tracklets. Since ArTIST relies solely on geometric information, on its own, it cannot estimate which of the $S$ inpainted options are valid.  
To select one of these candidates, we use a tracklet rejection scheme (TRS), as follows: if there is a candidate to be selected, we compute the intersection over union (IoU) of the last generated bounding box with all the detections in the scene. The model then selects the candidate with highest IoU, if it surpasses a threshold. However, in some cases, the last generated bounding box of one of the candidates may overlap with a false detection or a detection for another object, \textit{i.e.}, belonging to a different tracklet. To account for these ambiguities, we continue predicting  boxes for all candidates for another 1--2 frames and compute the IoUs for these frames as well. ArTIST then selects the candidate with the maximum sum of IoUs. This allows us to ignore candidates matching a false detection or a detection for another object moving in a different direction. However, this may not be enough to disambiguate all cases, \textit{e.g.}, the detections belonging to other tracklets that are close-by and moving in the same direction. We treat these cases in our assignment strategy discussed below. 

\subsection{Assignment}
\label{subsec:assignment}
To assign the detections to the tracklets at each time-step, we use the linear assignment found by the Munkres algorithm~\cite{munkres1957algorithms}. This method relies on a cost matrix $C$, storing the cost of assigning each detection to each tracklet. 
In our case, the costs are negative log-likelihoods computed by ArTIST. Let us denote by 
$C_{ij}^t = -\log p(\langle d_i^t,\mathcal{T}_j^t\rangle)$ the negative log-likelihood of assigning detection $i$ to tracklet $j$ at time $t$.
The Munkres algorithm then returns the indices of associated tracklet-detection pairs by solving 
$A^\star = \arg\min_{A^t} \sum_{i,j}C^t_{ij}A^t_{ij}$, where $A^t\in[0,1]^{N\times M}$ is the assignment probability matrix, with $N$ the number of detections and $M$ the number of tracklets. This matrix satisfies the constraints $\sum_j{A^t_{ij}=1},\;\forall i$ and $\sum_i{A^t_{ij}=1},\;\forall j$.

In practice, to account for the fact that we are less confident about the tracklets that we inpainted, we run the Munkres algorithm twice. First, using only the tracklets whose scores at the previous time-step were obtained using actual detections; second, using the remaining tracklets obtained by inpainting and the unassigned detections.

\section{Experiments}
\label{sec:experiment}
In this section, we evaluate different aspects of ArTIST and compare it with existing methods. 
In our experiments, bold and underlined numbers indicate the best and second best results, respectively. We provide the implementation details of our approach in the supplementary material. 
\\
\noindent\textbf{Datasets.} We use MOTChallenge benchmarks\footnote{\url{https://motchallenge.net/}}. 
MOTChallenge consists of several challenging pedestrian tracking sequences with moving and stationary cameras capturing the scene from various viewpoints and at different frame rates. We report our results on the three benchmarks of this challenge, MOT16~\cite{milan2016mot16}, MOT17~\cite{milan2016mot16}, and the recently introduced MOT20~\cite{dendorfer2020mot20}. MOT17 contains 7 training-testing sequence pairs with similar statistics. Three sets of public detections, namely DPM~\cite{felzenszwalb2009object}, Faster R-CNN~\cite{ren2015faster} and SDP~\cite{yang2016exploit}, are provided with the benchmark. The sequences of MOT16 are similar to those of MOT17, with detections computed only via DPM. 
MOT20 contains videos of very crowded scenes, in which there are many long occlusions occurring frequently. 
This dataset consists of 8 different sequences from 3 different scenes that are captured in both indoor and outdoor locations, during day and night. 
This dataset contains over 2M bounding boxes and 3,833 tracks, 10 times more than MOT16. 
For the ablation studies, we follow the standard practice of~\cite{zhou2020tracking} and thus split each training sequence into two halves, and use the first half for training and the rest for validation. Note that our main results are reported on the test sets of each benchmark dataset.
In all of our experiments, unless otherwise stated, we follow the standard practice of refining the public detections, which is allowed by the benchmarks and commonly done by the challenge participants~\cite{bergmann2019tracking,braso2020learning,karthik2020simple,xu2020train,zhou2020tracking}.
\\
\noindent\textbf{Evaluation Metrics.}
To evaluate MOT approaches, we use the standard metrics~\cite{bernardin2008evaluating,ristani2016performance} of MOT Accuracy (MOTA), Identity F1 Score (IDF1), number of identity switches (IDs), mostly tracked (MT), mostly lost (ML), false positives (FP), and false negatives (FN). The details of these metrics are provided in the supplementary material.

\subsection{Comparison with the State of the Art}
We compare ArTIST with existing approaches that, as ours, use the public detections provided by the benchmarks. 
In this section, we evaluate ArTIST in two settings, ArTIST-T, which utilizes the bounding box regression of~\cite{bergmann2019tracking}, and ArTIST-C, which uses the bounding box regression of~\cite{zhou2020tracking}, both acting on the public detections provided by the MOT benchmark datasets. For the sake of completeness, we consider both online and offline approaches. However, only online approaches are directly comparable to ArTIST. 

As clearly shown in our results in 
Tables~\ref{tab:motchallenge_mot16}, \ref{tab:motchallenge_mot17}, and~\ref{tab:motchallenge_mot20},
thanks to its stochastic, multi-modal motion model, ArTIST is capable of maintaining the identities of the tracklets for longer time periods, evidenced by
superior IDF1 scores.
Doing so allows ArTIST to keep more tracklets for more than 80\% of their actual lifespan, resulting in very high MT and very low IDs, outperforming all other competing methods. 
Another important aspect of ArTIST is its capability to inpaint the gaps due to detection failures. Filling such gaps not only has a great impact on identity preservation, but also significantly reduces the FN, a metric that is often ignored by existing trackers. \fs{As such, it directly affects the MOTA metric\footnote{MOTA is defined as $1-\sum_t(FN_t+FP_t+IDw_t) / \sum_tGT_t$.}, as there exist considerably  more  FN  than  FP and IDs, according to which our approach again outperforms existing methods by a considerable margin.}

As clearly evidenced by the performance of our approach on the challenging MOT20~\cite{dendorfer2020mot20} benchmark dataset, ArTIST is also a better tracker in highly crowded scenarios with frequent occlusions. In this benchmark, the mean crowd density is 10 times higher than in MOT16 and MOT17, reaching 246 pedestrians per frame. ArTIST's significant improvement in almost all MOT metrics demonstrates the benefit of using a better motion model, performing stochastic sampling for tracklet inpainting, and employing a probabilistic scoring function.

\begin{table}
    \centering
     \centering
     \caption{Results on MOT16 benchmark dataset on test set. The second column (RB) indicates whether the baselines use bounding box refinement.}
    \label{tab:motchallenge_mot16}
    \tabcolsep=0.08cm
    \scalebox{0.7}{
    \begin{tabular}{ l c  c  c c c c c c c }
    \toprule
    Method & RB & Mode  & MOTA $\uparrow$ & IDF1 $\uparrow$& IDs $\downarrow$ & MT $\uparrow$ &  ML $\downarrow$ & FP $\downarrow$ & FN $\downarrow$ \\
    \midrule
    JPDA~\cite{hamid2015joint}& \xmark & Off & 26.2 & - & 365 & 4.1 & 67.5 & 3,689 & 130,549 \\
    BiLSTM~\cite{kim2018multi}& \xmark & Off & 42.1 & 47.8 & 753 & 14.9 & 44.4 & 11,637 & 93,172 \\
     MHT-DAM~\cite{kim2015multiple}& \xmark & Off & 45.8 & 46.1 & 590 & 16.2 & 43.2 & 6,412 & 91,758 \\
    LMP~\cite{tang2017multiple}& \xmark & Off & 48.8 &  51.3 & 481 & 18.2 & 40.1 & 6,654 & 86,245\\
     MPNTrack~\cite{braso2020learning} & \cmark & Off & 58.6 & 61.7 & 354 & 27.3 & 34.0 & 4,949 & 70,252 \\
     \midrule
    EAMTT~\cite{sanchez2016online}& \xmark & On & 38.8 & 42.4 & 965 & 7.9 & 49.1 & 8,114 & 102,452  \\
    AMIR~\cite{sadeghian2017tracking}& \xmark & On & 47.2 & 46.3 & 774 & 14.0 & 41.6 & \underline{2,681} & 92,856 \\
    DMAN~\cite{zhu2018online}& \xmark & On & 46.1 & 54.8 & \underline{532} & 17.4 & 42.7 &   7,909 & 89,874 \\
    MOTDT~\cite{long2018real}& \xmark & On & 47.6 & 50.9 & 792 & 15.2 & 38.3 & 9,253 & 85,431\\
    STRN~\cite{xu2019spatial}& \xmark & On & 48.5 & 53.9 & 747 & 17.0 & 34.9 & 9,038 & 84,178 \\
    Tracktor++~\cite{bergmann2019tracking}& \cmark & On & 54.4 & 52.5 & 682 & 19.0 & 36.9 & 3,280 & 79,149 \\
    Tracktor++v2~\cite{bergmann2019tracking}& \cmark & On & 56.2 & 54.9 & 617 & 20.7 & 35.8 & \textbf{2,394} & 76,844\\
    DeepMOT-T~\cite{xu2020train}& \cmark & On & 54.8 & 53.4 & 645 & 19.1 & 37.0 &2,955 & 78,765 \\
    UMA~\cite{yin2020unified}& \cmark & On & 50.5 & 52.8 & 685 & 17.8 & \underline{33.7} & 7,587 & 81,924\\
    \midrule
    ArTIST-T & \cmark & On & \underline{56.6} & \underline{57.8} & \textbf{519} & \underline{22.4} & 37.5 & 3,532 & \underline{75,031}\\
    ArTIST-C  & \cmark & On & \textbf{63.0} & \textbf{61.9} & 635 & \textbf{29.1} & \textbf{33.2} & 7,420 & \textbf{59,376}\\
    \bottomrule
    \end{tabular}
    }
\end{table}

\begin{table}
    \centering
     \centering
     \caption{Results on MOT17 benchmark dataset on test set. The second column (RB) indicates whether the baselines use bounding box refinement. Note that CenterTrack*~\cite{zhou2020tracking} utilizes re-birthing.}
    \label{tab:motchallenge_mot17}
    \tabcolsep=0.08cm
    \scalebox{0.68}{
    \begin{tabular}{ l c  c  c c c c c c c }
    \toprule
    Method & RB & Mode  & MOTA $\uparrow$ & IDF1 $\uparrow$& IDs $\downarrow$ & MT $\uparrow$ &  ML $\downarrow$ & FP $\downarrow$ & FN $\downarrow$ \\
    \midrule
    IOU17~\cite{bochinski2017high} & \xmark & Off 
    & 45.5  & 39.4 & 5,988 & 15.7 & 40.5 & 19,993 & 281,643 \\
    EEB\&LM~\cite{maksai2019eliminating}& \xmark & Off 
    & 44.2 & 57.2 & 1,529 & 16.1 & 44.3 & 29,473 & 283,611 \\
    BiLSTM~\cite{kim2018multi}& \xmark & Off 
     & 47.5 & 51.9 & 2,069 & 18.2 & 41.7 & 25,981 & 268,042\\
    MHT-DAM~\cite{kim2015multiple}& \xmark& Off & 50.7 & 47.2 & 2,314 & 20.8 & 36.9 & 22,875 & 252,889  \\
    eHAF~\cite{sheng2018heterogeneous}& \xmark&Off&51.8 & 54.7 & 1,843 & 23.4 & 37.9 & 33,212 & 236,772 \\
    MPNTrack~\cite{braso2020learning} & \cmark & Off & 58.8 & 61.7 & 1,185 & 28.8 & 33.5 & 17,413 & 213,594\\
    \midrule
    MOTDT~\cite{long2018real}& \xmark & On & 50.9 & 52.7 & 2,474 & 17.5 & 35.7 & 24,069 & 250,768\\
    DMAN~\cite{zhu2018online}& \xmark & On & 48.2 & 55.7 & 2,194 & 19.3 & 38.3 & 26,218 & 263,608 \\
    EAMTT~\cite{sanchez2016online}& \xmark & On 
    & 42.6 & 41.8& 4,488 & 12.7 & 42.7 & 30,711 & 288,474 \\
    GMPHD~\cite{kutschbach2017sequential}& \xmark & On 
    & 39.6  & 36.6& 5,811 & 8.8 & 43.3 & 50,903 & 284,228\\
    SORT17~\cite{bewley2016simple} & \xmark & On 
     & 43.1 & 39.8& 4,852 & 12.5 & 42.3 & 28,398 & 287,582 \\
     STRN~\cite{xu2019spatial}& \xmark & On & 50.9 & 56.5 & 2,593 & 20.1 & 37.0 & 27,532 & 246,924 \\ 
    Tracktor++~\cite{bergmann2019tracking} & \cmark & On & 53.5 & 52.3 & 2,072 & 19.5 & 36.6 & 12,201 & 248,047 \\
    Tracktor++v2~\cite{bergmann2019tracking} & \cmark & On & 56.5 & 55.1 & 3,763 & 21.1 & 35.3 & \textbf{8,866} & 235,449\\
    DeepMOT-T~\cite{xu2020train}& \cmark & On & 53.7 & 53.8 & \underline{1,947} & 19.4 & 36.6 & \underline{11,731} & 247,447 \\
    FAMNet~\cite{chu2019famnet}& \cmark & On & 52.0 & 48.7 & 3,072 & 19.1 & 33.4 & 14,138 & 253,616\\
    UMA~\cite{yin2020unified}& \cmark & On & 53.1 & 54.4 & 2,251 & 21.5 & \underline{31.8} & 22,893 & 239,534\\ 
    CenterTrack*~\cite{zhou2020tracking} & \cmark & On & \underline{61.5} & \underline{59.6} & 2,583 & 26.4 & 31.9 & 14,076 & 200,672\\
    CenterTrack~\cite{zhou2020tracking} &\cmark & On & 61.4 & 53.3 & 5,326 & \underline{27.9} & \textbf{31.4} & 15,520 & \underline{196,886}\\
    \midrule
    ArTIST-T & \cmark & On & 56.7 & 57.5 & \textbf{1,756} & 22.7 & 37.2 & 12,353 & 230,437\\
    ArTIST-C & \cmark & On & \textbf{62.3} &\textbf{ 59.7} & 2,062 & \textbf{29.1} & 34.0 & 19,611 & \textbf{191,207}\\

    \bottomrule
    \end{tabular}
    } 
\end{table}

\begin{table}
    \centering
     \centering
     \caption{Results on MOT20 benchmark dataset on test set. Note that the methods denoted by * are the ones reported on CVPR2019 Challenge in which the videos are similar to MOT20 with very minor corrections in the ground-truth. The second column (RB) indicates whether the baselines use bounding box refinement.}
    \label{tab:motchallenge_mot20}
\tabcolsep=0.08cm
    \scalebox{0.68}{
    \begin{tabular}{ l c  c  c c c c c c c }
    \toprule
    Method & RB & Mode  & MOTA $\uparrow$ & IDF1 $\uparrow$& IDs $\downarrow$ & MT $\uparrow$ &  ML $\downarrow$ & FP $\downarrow$ & FN $\downarrow$ \\
    \midrule
    IOU19~\cite{bochinski2017high}* & \xmark & Off & 35.8 & 25.7 & 15,676 & 10.0 & 31.0 & 24,427 & 319,696\\
    V-IOU~\cite{bochinski2018extending}* & \xmark & Off & 46.7 & 46.0 & 2,589 & 22.9 & 24.4 & 33,776 & 261,964\\
    MPNTrack~\cite{braso2020learning} & \cmark & Off & 57.6	& 59.1 & 1,210 & 38.2 & 22.5 & 16,953 & 201,384 \\
    \midrule
    SORT20~\cite{bewley2016simple} & \xmark & On & 42.7 & 45.1 & 4,470 & 16.7 & 26.2 & 27,521 & 264,694\\
    TAMA~\cite{yoon2020online}* & \xmark & On & 47.6 & \underline{48.7} & \underline{2,437} & \underline{27.2} & \textbf{23.6} & 38,194 & \underline{252,934}\\
    Tracktor++~\cite{bergmann2019tracking}* & \cmark & On & \underline{51.3} & 47.6 & 2,584 & 24.9 & \underline{26.0} & \underline{16,263} & 253,680\\
    \midrule
    ArTIST-T & \cmark & On & \textbf{53.6} & \textbf{51.0} & \textbf{1,531} & \textbf{31.6} & 28.1 & \textbf{7,765} &\textbf{230,576} \\

    \bottomrule
    \end{tabular}
    } 
\end{table}

\subsection{Ablation Study}
\label{sec:ablation}
\begin{figure}
    \centering
    \includegraphics[width=0.48\textwidth]{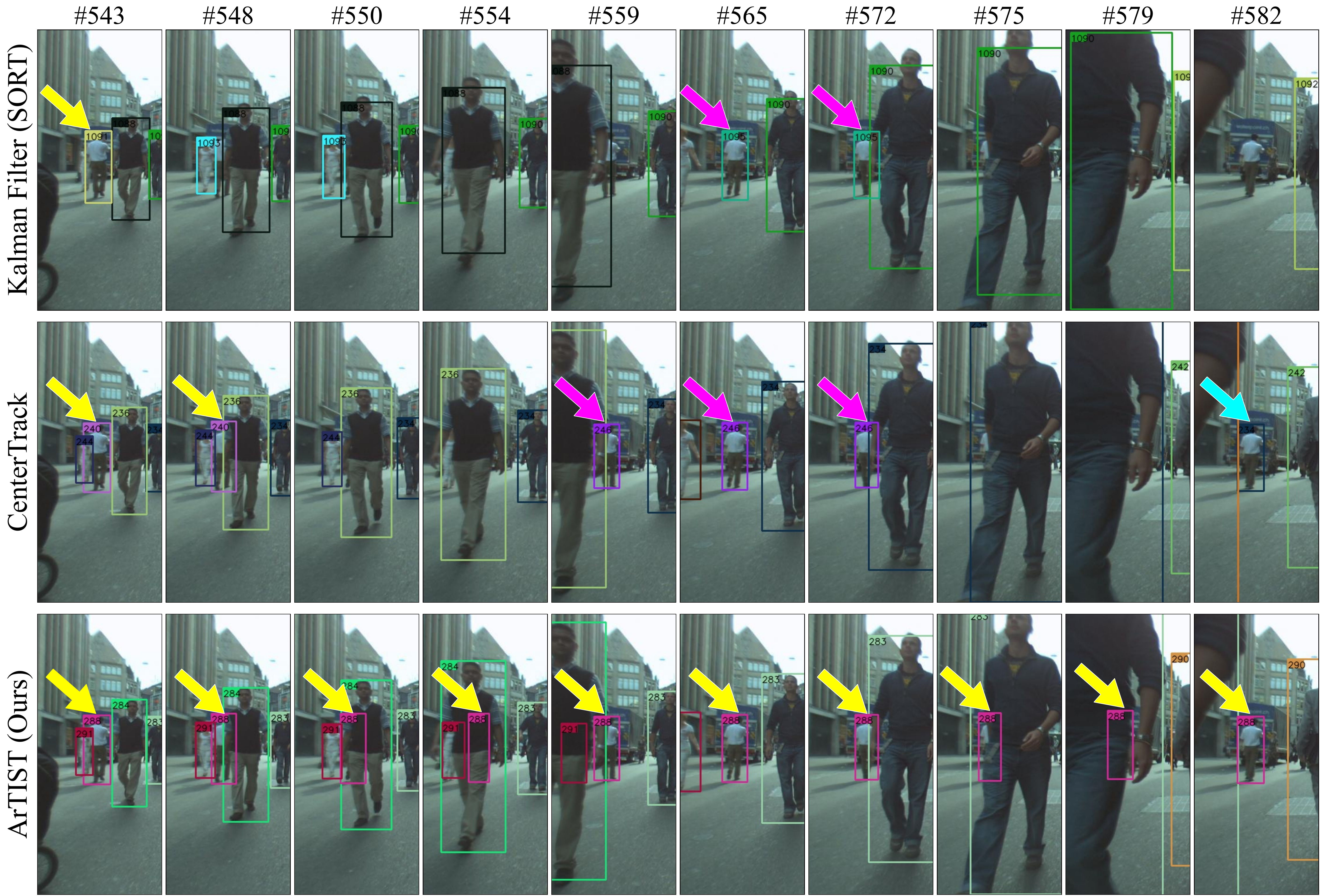}
    \caption{Qualitative result of handling occlusion in a moving camera scenario where colored arrows point to the bounding boxes of the pedestrian of interest and changing in the color of arrows shows a change in the identity of that pedestrian. Unlike Kalman Filter~\cite{bewley2016simple} and CenterTrack~\cite{zhou2020tracking}, our method preserves the identity after two occlusions and also inpaints the bounding boxes in occlusions. Note, all the methods are using exactly the same detections. }
    \vspace{-10pt}
    \label{fig:visual}
\end{figure}

In this section, we evaluate different components of ArTIST using ArTIST-C on the MOT17 validation set with the public Faster-RCNN~\cite{ren2015faster} detections.
\\
\noindent\textbf{Effect of Utilizing Interactions.}
Most existing trackers  treat individual tracklets independently of the other agents in the scene, ignoring the fact that the motion of each person is affected by that of the other pedestrians. 
This typically results in an increase of identity switches when different pedestrians are moving toward and passing each other, thus directly affecting the identity preservation capability of a tracker, which can be seen in the IDF1, IDs, and MT metrics. In Table~\ref{tab:interaction}, we evaluate the effect of our approach to accounting for interactions across agents, as described in Section~\ref{subsec:artist_arch}, by comparing it to the ``Isolation'' setting, where no interactions are utilized. Note that exploiting interactions improves almost all metrics, except for FP. 
\fs{A better identity preservation capability leads to an inevitable slight increase in FP since there are more attempts toward inpainting continuations of tracklets in case of occlusions, which is discussed below.}
\\
\noindent\textbf{Effect of Inpainting.}
As discussed in Section~\ref{sec:inpainting}, filling in the gap to compensate detector's failure leads to better identity preservation in a tracking framework. We demonstrate this effect in Table~\ref{tab:inpainting}, where we compare the no-inpainting case, with inpainting in visible or invisible mode. In the invisible mode, we do not consider the inpainted bounding boxes in the evaluations, whereas in the visible mode we do. As shown by our results, inpainting significantly improves the identity-sensitive metrics, such as IDF1, IDs, MT, and ML.
This experiments also shows that incorporating the inpainted bounding boxes (\textit{i.e.}, the visible mode) improves FN significantly \fs{which has a direct impact on MOTA.}
We observe that, while the inpainted tracklets resemble natural human motion, not all inpainted boxes correctly match the ground truth\footnote{This could also be due to the fact that GT provides an approximation of the position in case of occlusion.}, leading to a slight increase in FP and IDs. However, since FN is typically two to three orders of magnitude higher than FP and IDs, we see an overall improvement in tracking. In Fig.~\ref{fig:visual}, we provide a qualitative evaluation of the effect of inpainting, showing that our approach can better handle multiple occlusions.
\begin{table}[t]
    \centering
    \caption{Evaluating the effect of utilizing interactions.}
    \label{tab:interaction}
    \scalebox{0.72}{
    \begin{tabular}{l c c c c c c c}
    \toprule
    Setting & MOTA $\uparrow$ & IDF1$\uparrow$ & IDs$\downarrow$ & MT$\uparrow$ & ML$\downarrow$ & FP$\downarrow$ & FN$\downarrow$\\
    \midrule
    Isolation & 58.6 & 62.0 & 293 & 30.4 & 28.0 & \textbf{1,466} & 20,556\\
    Interaction & \textbf{59.8} & \textbf{66.5} & \textbf{231} & \textbf{33.3} & \textbf{27.7} & 1,675 & \textbf{19,769}\\
    \bottomrule
    \end{tabular}
    }
\end{table}
\begin{table}[t]
    \centering
    \caption{Evaluating the effect of tracklet inpainting.}
    \label{tab:inpainting}
    \scalebox{0.68}{
    \begin{tabular}{l  c c c c c c c}
    \toprule
    Setting & MOTA $\uparrow$ & IDF1$\uparrow$ & IDs$\downarrow$ & MT$\uparrow$ & ML$\downarrow$ & FP$\downarrow$ & FN$\downarrow$\\
    \midrule
    No Inpainting & 56.2 & 60.7 & 292 & 24.8 & 30.7 & \textbf{1,150} & 22,168\\
    Invisible Inpainting & 56.6 & 64.4 & \textbf{216} & 25.4 & 30.1 & 1,173 & 22,004\\
    Visible Inpainting &  \textbf{59.8} & \textbf{66.5} & 231 & \textbf{33.3} & \textbf{27.7} & 1,675 & \textbf{19,769}\\
    \bottomrule
    \end{tabular}
    }
\end{table}
\begin{table}[t]
    \centering
    \caption{Evaluating the effect  of multinomial sampling and TRS.}
    \label{tab:trs}
    \scalebox{0.71}{
    \begin{tabular}{l c c c c c c c}
    \toprule
    Setting & MOTA $\uparrow$ & IDF1$\uparrow$ & IDs$\downarrow$ & MT$\uparrow$ & ML$\downarrow$ & FP$\downarrow$ & FN$\downarrow$\\
    \midrule
    Top-1 & 58.8 & 62.5 & 293 & 30.7 & 28.0 & \textbf{1,444} & 20,484\\
    Multi. & 59.4 & 64.9 & 257 & 32.7 & 28.0 & 1,733 & 19,870\\
    Multi.+TRS &  \textbf{59.8} & \textbf{66.5} & \textbf{231} & \textbf{33.3} & \textbf{27.7} & 1,675 & \textbf{19,769}\\
    \bottomrule
    \end{tabular}
    }
\end{table}
\begin{table}[t]
    \centering
    \caption{Evaluating the effect of motion modeling. As opposed to other methods, our approach utilizes a multi-modal stochastic motion model.}
    \label{tab:motion}
    \scalebox{0.69}{
    \begin{tabular}{l c c c c c c c}
    \toprule
    Setting & MOTA $\uparrow$ & IDF1$\uparrow$ & IDs$\downarrow$ & MT$\uparrow$ & ML$\downarrow$ & FP$\downarrow$ & FN$\downarrow$\\
    \midrule
    No Motion & 58.1 & 60.7 & 614 & 31.3 & \textbf{25.6} & 1,572 & 20,379\\
    Kalman Filter & 56.8 & 61.9 & 247 & 24.7 & 30.7 & \textbf{1,185} & 21,865\\
    CenterTrack~\cite{zhou2020tracking} & 58.0 & 60.6 & 509 & 31.2 & 25.9 & 1,795 & 20,337\\
    Ours &  \textbf{59.8} & \textbf{66.5} & \textbf{231} & \textbf{33.3} & 27.7 & 1,675 & \textbf{19,769}\\
    \bottomrule
    \end{tabular}
    }
\end{table}
\begin{table}[t]
    \centering
    \caption{Evaluating the effect of public bounding box refinement.}
    \label{tab:refine}
    \scalebox{0.63}{
    \begin{tabular}{l c c c c c c c}
    \toprule
    Setting & MOTA $\uparrow$ & IDF1$\uparrow$ & IDs$\downarrow$ & MT$\uparrow$ & ML$\downarrow$ & FP$\downarrow$ & FN$\downarrow$\\
    \midrule
    Not Refined & 48.1 & 57.0 & 255 & 27.1 & 30.7 & 1,411 & 26,327\\
    ArTIST-C & 59.8 &66.5 & 231 & 33.3 & 27.7 & 1,675 & 19,769\\
    ArTIST-C (Private) & 68.5 & 71.1 & 229 & 42.2 & 21.5 & 2,343 & 14,409\\
    \bottomrule
    \end{tabular}
    }
\end{table}
\\
\noindent\textbf{Effect of Multinomial Sampling.} As discussed in Section~\ref{sec:inpainting}, 
ArTIST is capable of generating multiple plausible motion continuations by multinomial sampling from the learned  distribution.
In Table~\ref{tab:trs}, we compare a model that ignores the stochasticity in human motion, and thus greedily generates a single continuation of a tracklet for inpainting (denoted by ``Top-1''), with one that takes stochasticity into account (denoted by ``Multi.''). Note that, with more inpainted options, the model achieves better performance. However, large numbers of samples may introduce ambiguities in the system, causing a decrease in tracking performance. To handle this, we disambiguate such scenarios using our tracklet rejection strategy, whose results are provided in the third row of Table~\ref{tab:trs}. 
This experiment shows that, for sequences captured by a static camera, and for tracklets with relatively long observations, Top-1 sampling performs reasonably well, almost on par with multinomial sampling. This is due to the fact that, with long observations, our approach captures the motion pattern and can reliably fill in the gaps. However, when it comes to moving cameras or newly born tracklets (with relatively short observations), multinomial sampling (with tracklet rejection) leads to more reliable tracking.
\\
\noindent\textbf{Effect of Stochastic Motion Modeling.}
The key component of our approach is our stochastic motion model that is capable of capturing the multi-modal nature of human motion. To evaluate its effectiveness, given the same set of detections, we compare it with no motion model (setting CenterTrack's offsets~\cite{zhou2020tracking} to zero), a linear and uni-modal probabilistic motion model (Kalman Filter~\cite{bewley2016simple}), and a non-linear and deterministic motion model (existing state-of-the-art CenterTrack~\cite{zhou2020tracking}) in Table~\ref{tab:motion}. As shown by the results in the table and in Fig.~\ref{fig:visual}, the effect of learning a multi-modal distribution in scoring and inpainting is directly proportional to the success of the model at handling occlusions and thus at preserving identities for a longer time, resulting in a considerable improvement in metrics such as IDF1, IDs, and MT. 
\\
\noindent\textbf{Effect of Bounding Box Refinement.}
A number of recent tracking techniques~\cite{bergmann2019tracking,maksai2019eliminating,xu2020train,zhou2020tracking} refine the bounding boxes computed by the detectors. In particular,~\cite{bergmann2019tracking,xu2020train} use Faster R-CNN~\cite{ren2015faster} with ResNet-101~\cite{he2016deep} and Feature Pyramid Networks~\cite{lin2017feature} trained on  the MOT17Det~\cite{milan2016mot16} pedestrian detection dataset to refine the public detections provided with the MOTChallenge. Following~\cite{bergmann2019tracking}, CenterTrack~\cite{zhou2020tracking} also utilizes such refinement.
Note that, as acknowledged by~\cite{bergmann2019tracking, zhou2020tracking}, for the comparison with the methods that use the public detections to be fair, the new trajectories are still initialized from the public detection bounding boxes, and thus refinement is not used to detect a new bounding box. 
In this experiment, we evaluate the effectiveness of this refinement step. As shown by Table~\ref{tab:refine}, refinement leads to better tracking quality compared to the ``Not Refined'' setting, where the public detections are directly used in our tracking framework. Moreover, we evaluate the effect of using more accurate detected bounding boxes provided by a different detector, CenterNet~\cite{zhou2019objects}, which not surprisingly leads to even better tracking performance. 

\section{Conclusion}
We have introduced an online MOT framework based on a stochastic autoregressive motion model. Specifically, we have employed this model to both score tracklets for detection assignment purposes and inpaint tracklets to account for missing detections. 
Our results on the MOT benchmark datasets have shown the benefits of relying on a probabilistic multi-modal representation of motion, especially when dealing with challenging crowded scenarios with frequent occlusions, as in MOT20. Notably, without using any complex components, such as person Re-ID, our framework yields state of the art performance.

{\small
\bibliographystyle{ieee_fullname}
\bibliography{arxiv}
}

\newpage
\appendix

\section{Architecture Details}
As illustrated in Fig. 2 of the main paper, our novel model consists of two subnetworks, MA-Net and ArTIST. These two subnetworks are trained jointly. In this section, we introduce the architecture and implementation details of each of these subnetworks.

\paragraph{MA-Net.}
MA-Net is a recurrent autoencoder that is trained to capture the representation of motion of all agents in the scene. This is achieved by learning to reconstruct the motion of tracklets.
The subnetwork consists of an encoder that takes as input a 4D motion velocity representation, passes it through a fully-connected layer with 128 hidden units and a ReLU non-linearity, followed by a single GRU with 256 hidden units. The last hidden state of this (encoder) GRU initializes the hidden state of the decoder's GRU. The decoder is based on a residual GRU network that learns the velocity of changes in motion. To this end, given the initial hidden state and a seed 4D motion velocity representation (the velocity of changes between the first two consecutive frames), the decoder reconstructs each tracklet autoregressively.
On top of each GRU cell of the decoder, there exists a MLP that maps the hidden representation to a 4D output representation, \textit{i.e.}, the reconstructed velocity of motion at each time-step.

\paragraph{ArTIST.}
ArTIST takes as input a 4D motion velocity representation and a 256D interaction representation. The motion velocity is first mapped to a higher dimension via a residual MLP, resulting in a 512D representation. We then combine this with the interaction representation through concatenation. The resulting representation is then passed through a fully-connected layer that maps it to a 512D representation, followed by a ReLU non-linearity. This then acts as the input to a single layer LSTM with 512 hidden units to process the sequence. The LSTM produces a residual 512D vector, which is appended to its input to generate the final representation. To map the output of the LSTM to a probability distribution for each component of the motion velocity, we use 4 fully-connected layers (mapping 512D to \textit{K}D) followed by softmax activations, resulting in a $4\times K$ representation, where $K=1024$ is the number of clusters. 

\section{Implementation Details}
We train our model on a single GTX 2080Ti GPU with the Adam optimizer~\cite{kingma2014adam} for 110K iterations. We use a learning rate of 0.001 and a mini-batch size of 256. To avoid exploding gradients, we use the gradient-clipping technique of~\cite{pascanu2013difficulty} for all layers in the network. Since we use the ground-truth boxes during training, we apply random jitter to the boxes to simulate the noise produced by a detector. We train our model with sequences of arbitrary length (in range $[5, 100]$) in each mini-batch. During training, we use the teacher forcing technique of~\cite{williams1989learning}, in which ArTIST chooses with probability $P_{tf}$ whether to use its own output (a sampled bounding box) at the previous time-step or the ground-truth bounding box to compute the velocity at each time-step. We use $P_{tf}=0.2$ for the frames occurring after 70\% of the sequence length.
For our online tracking pipeline, we terminate a tracklet if it has not been observed for 30 frames. For tracklet rejection in the case of inpainting, we use an IOU threshold of 0.5 and set $t_{TRS}=1$ for low frame-rate videos and $t_{TRS}=2$ for high frame-rate ones. During multinomial sampling, we sample $\mathcal{S} = 50$ candidate tracklets. 
Note that, we also use the PathTrack~\cite{manen2017pathtrack} dataset, containing more than 15,000 person trajectories in 720 sequences, to augment MOT benchmark datasets. We implemented our model using the Pytorch framework of~\cite{paszke2017automatic}.

\begin{algorithm*}[!ht]
    \small
    \caption{ArTIST tracking at time $t$}\label{alg:artist_algo}
    \begin{algorithmic}[1]
        \Procedure{Tracking}{$D^t$, $\mathbb{T}$, $S$}
        \State Cost = \texttt{zeros}($|\mathbb{T}|$, $|D^t|$)
        \Comment{The cost matrix}
        \State FullSeq = [ ]
        \Comment{List of fully observed tracklets}
        \State InpaintSeq = [ ]
        \Comment{List of partially observed tracklets}
        \\
        \For{$\mathcal{T}_j$ in $\mathbb{T}$}
            \State $\Delta_j$ = \texttt{seq2vel}($\mathcal{T}_j$)
            \Comment{Compute motion velocity}
		    \State $I_j$ = \texttt{\bf agg}(MA-Net.\texttt{encode}($\mathbb{T} \setminus \{\mathcal{T}_j\}$))
		    \Comment{Compute interaction representation}
		    \\
		    \If{gap($\mathcal{T}_j$) == 0}
		    \Comment{Handle tracklets with full observation}
		        \State FullSeq.{\tt append}($j$)
		        \State $p(bbox_{\mathcal{T}_j}^t)$ = ArTIST($\Delta_j$, $I_j$)
		        \Comment{Compute the likelihood of next plausible bounding box}
		        
		        \For{$d_i^t$ in $D^t$}
		        \Comment{Compute the cost of assigning detections to $\mathcal{T}_j$}
		            \State Cost[$j$][$i$] = NLL($p(bbox_{\mathcal{T}_j}^t)$, $d_i^t$)
		        \EndFor
		        
		    \EndIf
		    \\
		    \If{gap($\mathcal{T}_j$) $>$ 0}
		    \Comment{Handle tracklets that require inpainting}
		        \State InpaintSeq.{\tt append}($j$)
		        \State $\widehat{\Delta}_{j, [1:S]}$ = {\tt Inpaint}($\Delta_j$, $I_j$, gap($\mathcal{T}_j$), $S$)
			    \Comment{Inpaint $S$ continuations for gap($\mathcal{T}_j$) time-steps}
			    \State $\widehat{\Delta}_j$ = \texttt{\bf TRS}($\widehat{\Delta}_{j, [1:S]}$, $D^t$)
			    \Comment{Choose the best of $S$ continuations}
			    \State $\Delta_j$ = [$\Delta_j$, $\widehat{\Delta}_j$]
			    \Comment{Synthesize full sequence}
			    \State $p(bbox_{\mathcal{T}_j}^t)$ = ArTIST($\Delta_j$, $I_j$)
			    \Comment{Compute the likelihood of next plausible bounding box}

		        \For{$d_i^t$ in $D^t$}
		        \Comment{Compute the cost of assigning detections to $\mathcal{T}_j$}
		            \State Cost[$j$][
		            $i$] = NLL($p(bbox_{\mathcal{T}_j}^t)$, $d_i^t$)
		        \EndFor
		        
		    \EndIf

        \EndFor
        \\
        \State $\text{assign}_f$, $\text{uD}_f$, $\text{uT}_f$ = {\tt Munkres}(Cost, $D^t$, $\mathbb{T}$, FullSeq)
        \Comment{Assignment for fully observed tracklets}
        
        \State $\text{assign}_i$, $\text{uD}_i$, $\text{uT}_i$ = {\tt Munkres}(Cost, $\text{uD}_f$, $\mathbb{T}$, InpaintSeq)
        \Comment{Assignment for inpainted tracklets (given the unassigned detections)}
	    \State assignment = {\tt Combine}($\text{assign}_f$, $\text{assign}_i$)
	    \Comment{Total assignment}
	    \State \textbf{update}($\mathbb{T}$, assignment)
	    \Comment{Update the tracklets at time-step $t$}
	    \EndProcedure
    \end{algorithmic}
\end{algorithm*}

\section{ArTIST Pseudo-code for Tracking}
In Algorithm~\ref{alg:artist_algo}, we provide the pseudo-code of our tracking algorithm. Following our discussion in Section 3 of the main paper, given the trained ArTIST model, detections, and current tracklets, this algorithm demonstrates how our approach updates tracklets at each time-step. 

\section{Evaluation Metrics}
Several metrics are commonly used to evaluate the quality of a tracking system~\cite{ristani2016performance,bernardin2008evaluating}. The main one is MOTA, which combines quantification of three error sources: false positives, false negatives and identity switches. A higher MOTA score implies better performance. Another important metric is IDF1, i.e., the ratio of correctly identified detections over the average number of ground-truth and computed detections. The number of identity switches, IDs, is also frequently reported. Furthermore, the following metrics provide finer details on the performance of a tracking system: mostly tracked (MT) and mostly lost (ML), that are respectively the ratio of ground-truth trajectories that are covered/lost by the tracker for at least 80\% of their respective life span; False positives (FP) and false negatives (FN). 
All  metrics were computed using the official evaluation code provided by the MOTChallenge benchmark.

\end{document}